\documentclass{article}
\usepackage{spconf}

\usepackage{times}
\usepackage{epsfig}
\usepackage{graphicx}
\usepackage{amsmath}
\usepackage{amssymb}
\usepackage{color}
\usepackage[table]{xcolor}
\usepackage[lofdepth,lotdepth, caption=false,font=footnotesize]{subfig}
\usepackage{hhline}
\usepackage{cite}
\usepackage{stfloats}
\definecolor{Gray}{gray}{0.85}

\newcommand{\VAEfig}{
\begin{figure*}[t]
    \centering
    \includegraphics[width=0.75\textwidth]{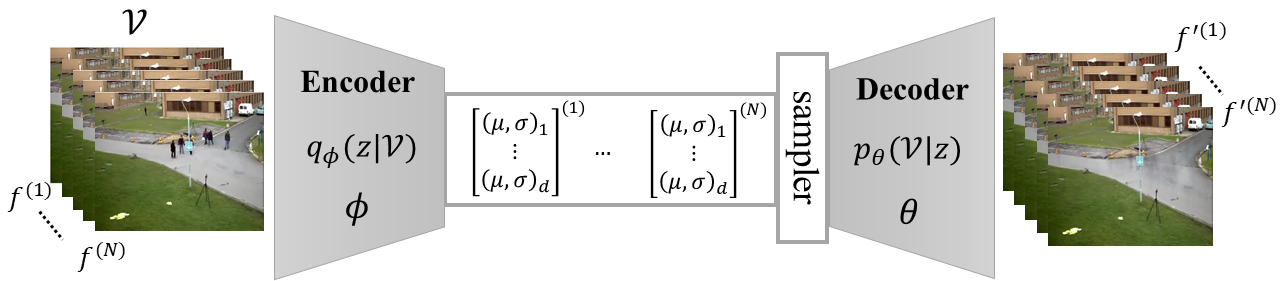}
    \caption{ \footnotesize Schematic of a variational autoencoder (VAE) for background subtraction used in our DeepPBM.}
    \label{fig:VAE}
    \vspace{-.15in}
\end{figure*}

}
\newcommand{\BMadapt}{
\begin{figure}[t]
    \centering
    \includegraphics[width=1\linewidth, trim=1.4in 1.5in 1.4in 1.5in, clip=true ]{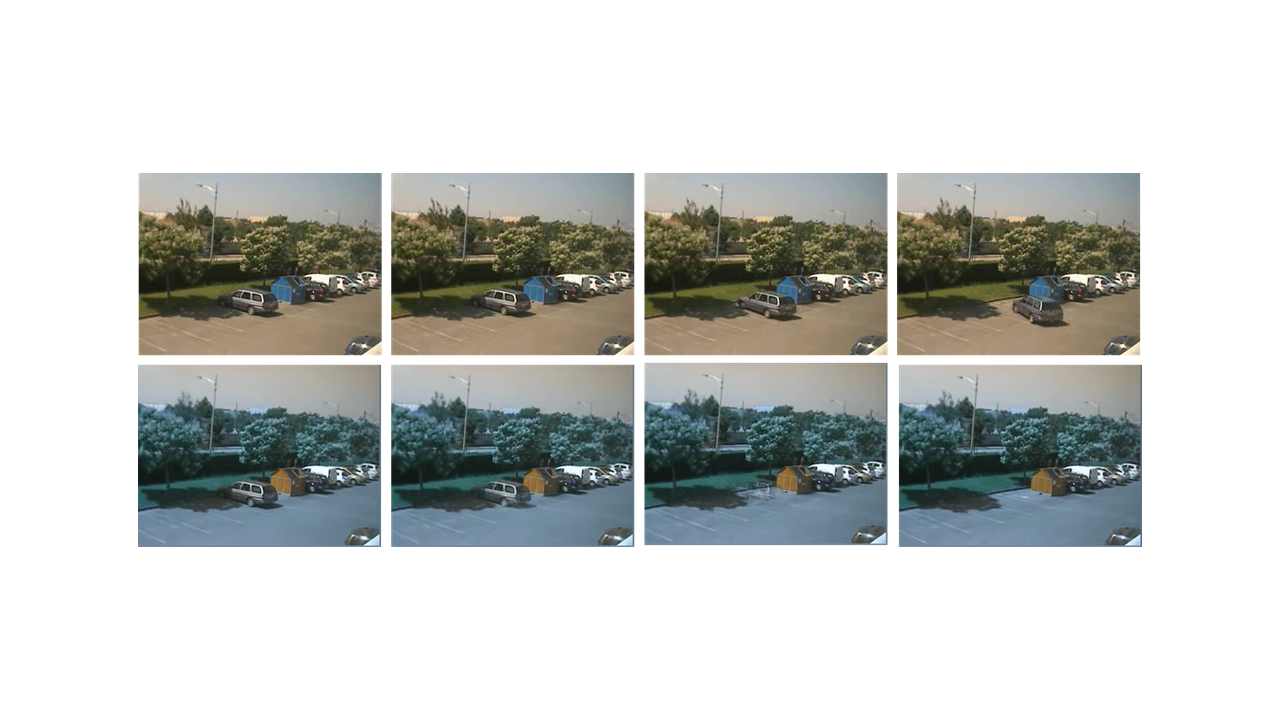}
    \caption{\footnotesize Adaptation of the background model estimated by DeepPBM to the changes of the scene for one of the long videos in BMC2012 dataset. First row shows the consecutive frames of the original video, second row shows the corresponding background model.}
    \label{fig:bmadapt}
    \vspace{-.15in}
\end{figure}
}

\newcommand{\BMCresult}{
\begin{figure}[t]
    \centering
    \includegraphics[width=1\linewidth, trim=0.1in 0.5in 0.1in 0.5in, clip=true ]{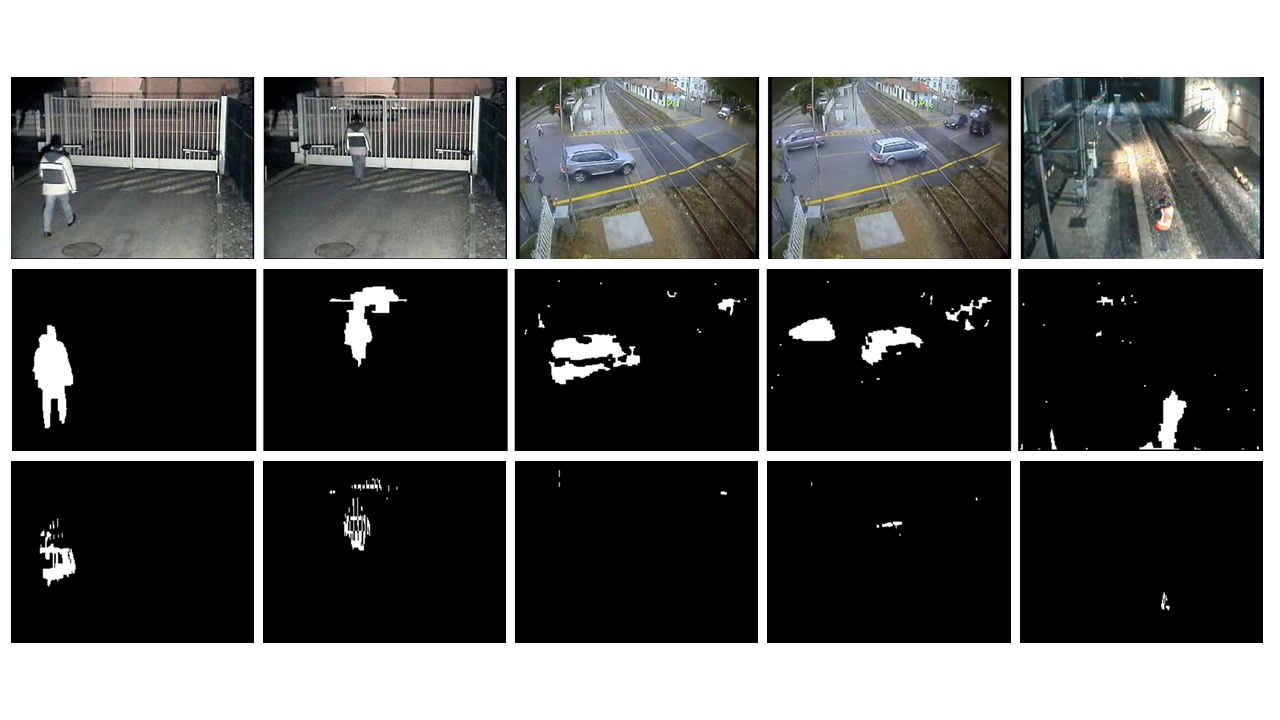}
    \caption{\footnotesize Extracted masks from estimated background model for some of the frames selected from sample videos in the BMC2012 dataset. First row shows the original frames, second row shows the corresponding masks resulted from DeepPBM, and third row shows the masks resulted from the RPCA method.}
    \label{fig:bmcresult}
    \vspace{-.2in}
\end{figure}
}
\newcommand{\tableresultBMC}{
\begin{table}[t] 
\scriptsize
\centering
\caption{ \footnotesize Benchmark metrics and execution time for the BS task of our DeepPBM compared to RPCA evaluated on the 6 short videos of BMC2012 dataset. For the fair comparison we ran the trained model on the CPU mode.}
\begin{center}  
 \begin{tabular}{| c | c | c | c | c|} 
 \hline
   Algorithm  & F-measure & Recall & Precision & Run Time \\ 
 \hline
  \rowcolor[gray]{0.8}
 \multicolumn{5}{|c|} {Big trucks -- $1498 ~\text{frames}$}   \\
 \hline
 \text{RPCA} &  0.68  & 0.6 & 0.80 & 18 min\\ 
 \hline
\text{DeepPBM ($d=30$)}  & 0.86 & 0.85 & 0.88 & 2.8 min\\ 
 \hline
   \rowcolor[gray]{0.8}
 \multicolumn{5}{|c|}{Wandering students -- $795 ~\text{frames}$} \\
 \hline
 \text{RPCA}  &  0.87 & 0.84 & 0.90 & 6.2 min\\
 \hline
 \text{DeepPBM ($d=20$)} &  0.94 & 0.92  & 0.95 & 1.1 min\\ 
 \hline
   \rowcolor[gray]{0.8}
 \multicolumn{5}{|c|}{Rabbit in the night -- $1896 ~ \text{frames}$} \\
 \hline
 \text{RPCA} & 0.60 & 0.59 & 0.61 & 28 min\\ 
 \hline
 \text{DeepPBM ($d=35$)}  &  0.90 & 0.94 & 0.87 & 2.7 min\\
 \hline
   \rowcolor[gray]{0.8}
 \multicolumn{5}{|c|}{Beware of the trains -- $1065 ~ \text{frames}$} \\
 \hline
 \text{RPCA} & 0.68 & 0.61 & 0.78 & 11.5 min\\
 \hline
 \text{DeepPBM ($d=30$)} & 0.81 & 0.83 & 0.78 & 1.5 min\\
\hline
  \rowcolor[gray]{0.8}
\multicolumn{5}{|c|}{Train in the tunnel -- $1726 ~ \text{frames}$} \\
 \hline
 RPCA  & 0.63 & 0.60 & 0.81  & 14.2 min\\
 \hline
 \text{DeepPBM ($d=30$)}  & 0.70 & 0.70  & 0.71 & 2.4 min\\
\hline
\rowcolor[gray]{0.8}
\multicolumn{5}{|c|}{Traffic during windy day -- $793~ \text{frames}$} \\
 \hline
 RPCA  &0.54  & 0.50 & 0.58 & 8.4 min\\
 \hline
 \text{DeepPBM ($d=1$)}  & 0.76 & 0.74  & 0.79 &  1.1 min\\
\hline
\rowcolor[gray]{0.9}
\multicolumn{5}{|c|}{\textbf{Average over all the videos}} \\
 \hline
 RPCA  & 0.67 & 0.62  &  0.75 &  14.4 min\\
 \hline
 \text{DeepPBM}  & \textbf{0.83} & \textbf{0.83}  &  \textbf{0.83} & \textbf{1.9 min}\\
\hline
 \end{tabular}
 \label{tbl:BMCresults}
\end{center}
\vspace{-0.3in}
\end{table}
}

\newcommand{\tableresultBMCLong}{
\begin{table}[t] 
\scriptsize
\centering
\caption{\footnotesize Benchmark metrics for the BS task of our DeepPBM on 3 long videos of BMC2012 dataset.}
\begin{center}  
 \begin{tabular}{| c | c | c | c | c |} 
 \hline
   Algorithm  & F-measure & Recall & Precision & Run Time\\ 
 \hline
  \rowcolor[gray]{0.8}
 \multicolumn{5}{|c|} {Video\_001 -- $22 ~ \text{min}$}   \\
 \hline
\text{DeepPBM ($d=30$)}  & 0.73 &  0.76 & 0.71 & 4.69 min\\ 
 \hline
   \rowcolor[gray]{0.8}
 \multicolumn{5}{|c|}{Video\_005 -- $78 ~ \text{min}$} \\
 \hline
 \text{DeepPBM ($d=30$)} & 0.71  & 0.73  &  0.62 & 16.67 min\\ 
 \hline
   \rowcolor[gray]{0.8}
 \multicolumn{5}{|c|}{Video\_009 -- $72 ~ \text{min}$} \\
 \hline
 \text{DeepPBM ($d=20$)}  & 0.63  & 0.70  & 0.68 & 15.34 min\\
 \hline
    \rowcolor[gray]{0.9}
 \multicolumn{5}{|c|}{\textbf{Average over all the videos}} \\
 \hline
 \text{DeepPBM}  &0.69  & 0.73  &  0.67 & 12.23 min\\
\hline
\end{tabular}
\label{tbl:resultBMCLong}
\end{center}
\vspace{-0.3in}
\end{table}
}

\newcommand{\tableNetArch}{
\begin{table}[t] 
\scriptsize
\centering
\caption{\footnotesize DeepPBM network architecture.}
\begin{center}
 \begin{tabular}{| c | c |} 
 \hline
 \rowcolor[gray]{0.8}
 \textbf{Layer \#} & \textbf{Encoder}    \\
 \hline
  & \text{Input: $w\times h\times 3$ RGB image}  \\ 
 \hline
 \text{1} & \text{$4\times 4$ conv, 32 Relu, stride 2, BatchNorm}   \\
 \hline
 \text{2} & \text{$4\times 4$ conv, 64 Relu, stride 2, BatchNorm} \\ 
 \hline
 \text{3} & \text{$4\times 4$ conv, 128 Relu, stride 2, BatchNorm}   \\
 \hline
 \text{4} & \text{$4\times 4$ conv, 128 Relu, stride 2, BatchNorm}  \\
 \hline
  & \text{Intermediate output: $128\times w^\prime\times h^\prime$ patch}  \\
 \hline
 \text{5} & \text{FC 2400 ReLU, Dropout 0.3} \\
 \hline
 \text{6} & \text{FC $2\times d$}  \\
 \hline
  & \text{Output: $\mu_z,\sigma^2_z\in\mathbb{R}^d$} \\
 \hline
 \hline
 \rowcolor[gray]{0.8}
 & \textbf{Decoder} \\
 \hline
 & \text{Input: $z\in\mathbb{R}^d$} \\
 \hline
 \text{1}& \text{FC 2400 ReLU} \\
 \hline
 \text{2}& \text{FC $128\times w^\prime\times h^\prime$ ReLU, Dropout 0.3} \\
 \hline
 \text{3}& \text{$4\times 4$ deconv, 128 Relu, stride 2, BatchNorm} \\
 \hline
 \text{4}& \text{$4\times 4$ deconv, 64 Relu, stride 2, BatchNorm} \\
 \hline
 \text{5}& \text{$4\times 4$ deconv, 32 Relu, stride 2, BatchNorm} \\
 \hline
 \text{6}& {$4\times 4$ deconv, 3 Sigmoid, stride 2} \\
 \hline
 & \text{Output: $w\times h\times 3$ RGB image} \\
 \hline
 \end{tabular}
 \label{tbl:NetArch}
\end{center}
\vspace{-0.3in}
\end{table}
}

\newcommand{\eqnref}[1]{Eq.~(\ref{eqn:#1})}
\newcommand{\figref}[1]{Fig.~\ref{fig:#1}}
\newcommand{\tblref}[1]{Table~\ref{tbl:#1}}
\newcommand{\secref}[1]{Section~\ref{sec:#1}}
\usepackage{hyperref}
\hypersetup{
    colorlinks=true,
    linkcolor=green,
    filecolor=magenta,      
    urlcolor=blue,
}

\begin{document}

\title{DeepPBM: Deep Probabilistic Background Model Estimation \\from Video Sequences}

\name{Amirreza Farnoosh*, Behnaz Rezaei*, and Sarah Ostadabbas
                      \thanks{* indicates equal contribution. Source code available at: \url{https://github.com/ostadabbas/DeepPBM}}}
                 \address{Augmented Cognition Lab (ACLab)\\
Northeastern University, Boston, MA, USA\\
\tt\small \{afarnoosh;brezaei,ostadabbas\}@ece.neu.edu}

\maketitle

\begin{abstract}

This paper presents a novel unsupervised probabilistic model estimation of visual background in video sequences using a variational autoencoder framework. Due to the redundant nature of the backgrounds in surveillance videos, visual information of the  background can be compressed into a low-dimensional subspace in the encoder part of the variational autoencoder, while the highly variant information of its moving foreground gets filtered throughout its encoding-decoding process. Our deep probabilistic background model (DeepPBM) estimation approach is enabled by the power of deep neural networks in learning compressed representations of video frames and reconstructing them back to the original domain. We evaluated the performance of our DeepPBM in background subtraction on 9 surveillance videos from the background model challenge (BMC2012) dataset, and compared that with a standard subspace learning technique, robust principle component analysis (RPCA), which similarly estimates a deterministic low dimensional representation of the background in videos and is widely used for this application. Our method outperforms RPCA on BMC2012 dataset with 23\% in average in F-measure score, emphasizing that background subtraction using the trained model can be done in more than 10 times faster. 
   
\end{abstract}

\section{Introduction}
Detection of moving objects or change detection in videos recorded can be seen as the process of separating the foreground from background. This process is a central component in every video surveillance, security, and traffic monitoring system. A huge body of research exists in the background vs. foreground separation topic since the introduction of simple yet effective mixture of Gaussian (MoG) model by Stauffer et al.\cite{stauffer1999adaptive}. Yet, development of an efficient background subtraction (BS) process for robust moving object detection that addresses the key challenges in dynamic backgrounds is not completely resolved. A competent BS algorithm should be fast and robust to the dynamic nature of the background. Furthermore, it should be implemented in an unsupervised manner to be able to generalized to the new scenes. Although several state-of-the-art algorithms have been proposed for adaptive background representation \cite{Biancohowfar2017, st2015subsense, allebosch2015efic, 7045991}, a universal method that can address different BS challenges present in long-term videos is still missing.
Recently, deep learning approaches based around using convolutional neural networks (CNNs) have shown promising results in BS problems in different challenging conditions \cite{lim2018foreground, babaee2018deep, wang2017interactive, Sakkos2017}. However, all of these methods are supervised and were trained on ground truth video frames of benchmark datasets and tested on the same types of videos. In addition, non of these BS algorithms has been evaluated on long-term videos to demonstrate their adaptation performance in real-world applications.

To provide an unsupervised, generalizable and computationally efficient solution to the problem of BS, we introduce a novel deep probabilistic background model (DeepPBM) estimation approach, which capitalizes on the power and flexibility of deep neural networks in approximating complex functions. Our approach is centered around two following hypotheses: (1) background in videos recorded by an stationary camera lies on a low-dimensional subspace represented by a series of latent variables, and (2) there is a Gaussian distribution model for the latent subspace of the background embedded by a non-linear mapping of the video frames. An important property of our DeepPBM approach is its generative modeling of the background, which can be used for creating synthetic backgrounds of the specific scene with different illuminations, shadings, and waving by variations in its latent variables. These synthetic backgrounds may be used for training purposes in deep learning models. The proposed DeepPBM shows high performance in BS in the majority of the scenes in the BMC2012 dataset \cite{vacavant2012benchmark}. DeepPBM is also observed to have an acceptable performance in adapting the background model in long-term videos in the this dataset performing orders of magnitude faster than its non-deep counterpart robust principle component analysis (RPCA).
\subsection{Overview of Background Subtraction Techniques}
\VAEfig
\label{sec:overview}
BS is usually achieved by creating a background model from video frame sequences, choosing a strategy to update this background model, and then subtracting each frame from this background model. The performance of the BS algorithm depends on how well each of theses steps can be implemented. After MoG presented in \cite{stauffer1999adaptive}, various follow-up works in probabilistic background modeling have been proposed to improve the performance of the MoG approach through employing different learning methods and adaptation modifications \cite{chen2007efficient,yong2013improved, haines2014background, kaewtrakulpong2002improved}. Nonetheless, these methods all suffer from the noise in the initial frames as well as inflexibility to the the sudden changes in the background throughout the video.

In parallel, a significant amount of research effort has been dedicated to the modeling of the video backgrounds as a low-dimensional subspace in the original high-dimensional space of the video frames. Considering this assumption, the problem of BS has been formulated as an optimization problem in different works, in which an observation video matrix is decomposed into a low rank matrix forming background sequence and an additive part representing the moving object as the foreground  \cite{candes2011robust,he2012incremental,zhou2013moving, he2012incremental, 6854862, rezaei2017background, 8456638}. Although these algorithms work visually well in modeling background and its gradual changes, they are constructed based on an optimization problem with heavy structural properties that requires to be solved by computationally expensive iterations, making them impractical for online video inspection applications.

Recently, there has been few efforts to employ the capability of deep neural networks (DNNs) in performing BS \cite{braham2016deep, wang2017interactive, lim2018foreground, babaee2018deep}. However, all of these approaches are performed in a supervised manner, and therefore require manual foreground mask extraction from a subset of video frames for their learning phase. In these works, authors train a specialized CNN to either find a supervised model of the background in video frames from a subset of manually annotated frames \cite{wang2017interactive} or finding foreground mask by doing the subtraction phase being provided by the background model from another method \cite{braham2016deep, babaee2018deep}. In \cite{lim2018foreground} authors proposed a triplet CNN with weak supervision for a multistage background feature embedding using an encoder-decoder structure. As we mentioned before, despite the high performance of current deep learning methods in foreground/ background segmentation, these methods are supervised and highly dependant on the quality of the background model that they use for BS. Moreover none of them are tested on long videos to show their adaptation quality in real applications that need long-term video inspection.
\vspace{-0.2in}
\section{Proposed DeepPBM Estimation Approach}
\label{sec:approach}
Variational autoencoders (VAEs) have emerged as one of the most popular approaches in unsupervised learning of complicated distributions that underlie models or generate data \cite{2013arXiv1312.6114K,2016arXiv160605908D}. VAEs are compelling since they can be set up in the framework of deep learning (DL), and therefore benefit from the ongoing advances in this field. In the context of DL, a VAE consists of an encoder and a decoder. 
Illustrated in \figref{VAE}, encoder learns an efficient representation of its input data and projects that into a stochastic lower dimensional space, determined by latent variables. The decoder tries to recover the original data, given the probabilistic latent variables from the encoder. The entire network is trained by comparing the original input data with its reconstructed output \cite{2016arXiv160605908D}. We further discuss the mathematical details of each part in \secref{VAEmodel} 

From an information theoretic perspective, the compression of the high-dimensional input to a low-dimensional space as done in the encoder part of VAE, and then decompressing it back to the original space leads to the loss of high variant information ( in our case moving objects), which is measured and used to learn the network. This lossy low-dimensional representation of the input data is a desired attribute that can be utilized in the context of BS in surveillance videos. This attribute follows similar principles employed in low-rank subspace learning approaches for unsupervised BS. Further, it can benefit from the power and flexibility of DL in learning a more effective low-dimensional space. Moreover, using DL allows us to transfer the computational cost of solving the subspace learning from the evaluation to the training process of DL, which could entirely be performed offline. Following aforementioned significance, the main idea behind our proposed DeepPBM is using VAE built on top of a DNN for the purpose of unsupervised BS considering the low-dimensional representation attribute of VAE along with the compression capacity of background images.
\subsection{Probabilistic Modeling of the Background in Videos}
\label{sec:VAEmodel}
Considering that video frames $f^{(i)}\in \mathcal{V}, i\in\{1,\dots,N\}$, each of size $w\times h$ pixels, are generated from $d$ underlying probabilistic latent variables vectorized in $z\in \mathbb{R}^d$ in which $d \ll w\times h$, the vector $z$ is interpreted as the compressed representation of the video. A VAE considers the joint probability of the input video, $\mathcal{V}$, and its representation, $z$, to define the underlying generative model as $p_\theta(\mathcal{V}, z) = p_\theta(\mathcal{V} \vert z) p(z)$, where $p(z) = \mathcal{N}(0, I)$ is the standard Gaussian prior for latent variables $z$, and $p_\theta(\mathcal{V}|z)$ is the decoder part of a VAE that is parameterized by a DNN with parameters $\theta$. In the encoder part of the VAE, the posterior distribution $p(z\vert\mathcal{V})$ is approximated with a variational posterior $q_\phi(z\vert\mathcal{V})$ with parameters $\phi$. Each dimension of the latent space in this variational posterior is modeled independently with a Gaussian mean and variance for each video frame, as $ q_\phi(z\vert f) =\prod_{k=1}^{d}\mathcal{N}(z_k\vert{\mu_k^f},{\sigma^f_k}^2)$, where $\mu^f$, and ${\sigma^f}^2$ are outputs of the encoder, $q_\phi(z\vert f)$, which is also parameterized by a DNN with parameters $\phi$.
The efforts in making this variational posterior as close as possible to the true posterior distribution results in maximization of the evidence lower bound (ELBO) \cite{2013arXiv1312.6114K,2016arXiv160100670B}, such that the final VAE objective for the entire video becomes:

\begin{align} \label{eqn:vae}
\scriptsize
    & ELBO_\mathcal{V} (\theta, \phi) = \\ \nonumber
    & \frac{1}{N}\sum_{i=1}^{N}\Big[\mathbb{E}_{q_\phi(z|f^{(i)})}\big[\log p_\theta(f^{(i)}|z)\big]- KL\big(q_\phi(z|f^{(i)})||p(z)\big)\Big]
\end{align}
The first term in \eqnref{vae} (expected likelihood term) can be interpreted as the negative reconstruction error, which encourages the decoder to learn to reconstruct the original input, and the second term is the Kullback-Leibler (KL) divergence between prior and variational posterior distribution of latent variables, which acts a regularizer to penalize the model complexity.

For our purpose of BS, we used an $l_1$-norm loss function for reconstruction error of the VAE in order to capture the sparsity of the foreground assumed in the majority of low rank subspace factorization studies used in background/foreground separation. The KL term can also be calculated analytically in the case of Gaussian distributions. Therefore, the total loss function for our proposed DeepPBM becomes:
\begin{align}
\scriptsize
\label{eqn:vae1}
 & Loss(f,f^\prime,\mu^f,{\sigma^f}^2)=  \\ \nonumber
 & \sum_{i=1}^{N}\vert f^{(i)}-f^{\prime{(i)}}\vert - \frac{1}{2} \sum_{i=1}^{N}\big( 1+\log{{\sigma^{f^{(i)}}}^2}-{\mu^{f^{(i)}}}^2-{\sigma^{f^{(i)}}}^2\big)
\end{align}
Where $f^\prime$ is the reconstructed version of the input video frame, $f$, produced by the decoder  \cite{2013arXiv1312.6114K,2016arXiv160605908D}. 

\tableNetArch
\subsection{DeepPBM Architecture and Training}
\label{sec:architecture}
The encoder and decoder parts of the VAE in the DeepPBM are both implemented using a CNN architecture specified in \tblref{NetArch}. The encoder takes the video frames as input and outputs the mean and variance of their underlying low dimensional latent variables distributions. The decoder takes samples drawn from latent distributions as input and output the recovered version of the original input. The network is trained by minimizing the error defined in \eqnref{vae1}. We trained the VAE using the gradient descent to optimize this loss with respect to the parameters of the encoder and decoder, $\theta$ and $\phi$, respectively. The input video data is trained in batches of size 140 for 200 epochs.

\tableresultBMC

\section{Performance Assessment}
\label{sec:experiment}
We evaluated the performance of our proposed algorithm, DeepPBM, in BS on the BMC2012 benchmark dataset \cite{vacavant2012benchmark}. This benchmark contains 9 real world surveillance videos along with encrypted ground-truth masks of the foreground for a subset of frames in each video. This dataset focuses on outdoor situations with various weather and illumination conditions such as wind, sun, or rain. Therefore, makes it suitable for performance evaluation of BS methods in challenging conditions. We used the short videos in this dataset to compare the estimation quality of DeepPBM against RPCA. We then used the long videos to examine how our DeepPBM adapts to changes in the background model over a long period of time. Please note that due to the shortage of the memory and processing units required for running RPCA, we could not apply RPCA for the long videos. The evaluation metrics are computed by the software that is provided with the dataset, based on the encrypted ground-truth masks.
In order to extract the masks of the moving objects in short videos (with less than 2000 frames), we first trained the DeepPBM network using all of the video frames as explained in \secref{architecture}. The dimension of the latent variables, $d$, needs to be tuned based on the dynamics/complexity of the background model in each video. For videos with dynamic background (e.g. in windy, rainy or snowy conditions), a larger $d$ should be selected in order to capture variations in the background, however, for videos with monotonic background (with slight or no changes in background along video frames) a smaller $d$ should be selected to prevent network from learning foreground. After the network was trained, we fed the same frames to the network to estimate the background image for each individual frame. Finally, we used the estimated background of each frame to find the mask of the moving objects by thresholding the difference between the original input frame and the estimated background. The quantitative results of the performance of DeepPBM in BS compared to the RPCA is reported in \tblref{BMCresults}. As it is observed, DeepPBM outperforms RPCA in all of the short videos by $23\%$ in F-measure. Further it performs more than 10 times faster than RPCA. \figref{bmcresult} illustrates sample results of applying DeepPBM and RPCA on short videos of BMC2012 dataset. As seen, DeepPBM is quite successful in detecting moving objects in these scenes, and generates acceptable masks of the foreground, while RPCA fails to detect accurate foreground masks.
\BMCresult
For the long videos, we used the first 20\% of the video frames for training of the DeepPBM, and then used this trained network to extract background images for all of the frames. \tblref{resultBMCLong} shows the quantitative performance of DeepPBM for the long videos which gives an average F-measure score of $0.69$. \figref{bmadapt} illustrates how the network adapts to the changes in the background that happen over a long period of time. The car in this sample scene is initially included as part of the background model in the first two frames, since it has been stationary for a long period, however, the network begins to detect that as foreground in the next two frames as soon as the car starts to move. 

\tableresultBMCLong

\BMadapt
\vspace{-.1in}
\section{Conclusion}
\label{sec:conclusion}
In this paper, we presented our DeepPBM method using the framework of VAE for detecting the moving objects in videos recorded by stationary cameras. We evaluated the performance of our model in the task of background subtraction, and showed how well it adapts to the changes of the background in long-term monitoring on the BMC2012 dataset. According to the reported results, DeepPBM outperformed RPCA known as one of the standard and well-performed subspace learning methods for background modeling in both time efficiency and modeling performance.
Note that our approach estimates a generative low-dimensional model of the background and task of the BS is performed by simply thresholding the difference between this model and the original input frame. One of the important directions in our future work will be  performing selective background updates via adapting the background model to the pixels that were detected as background by the network, as opposed to the current network fine-tuning paradigm after specific time intervals.

{\small
\bibliographystyle{ieee}
\bibliography{paper}
}
\end{document}